\documentclass[letterpaper]{article} 
\usepackage{aaai23}  
\usepackage{times}  
\usepackage{helvet}  
\usepackage{courier}  
\usepackage[hyphens]{url}  
\usepackage{graphicx} 
\usepackage{natbib}  
\usepackage{caption} 
\usepackage{algorithm}
\usepackage{algorithmic}

\usepackage{newfloat}
\usepackage{listings}
\DeclareCaptionStyle{ruled}{labelfont=normalfont,labelsep=colon,strut=off} 
\floatstyle{ruled}
\newfloat{listing}{tb}{lst}{}
\floatname{listing}{Listing}

\usepackage[utf8]{inputenc} 
\usepackage[T1]{fontenc}    
\usepackage{url}            
\usepackage{booktabs}       
\usepackage{amsfonts}       
\usepackage{nicefrac}       
\usepackage{microtype}      
\usepackage{graphicx}
\usepackage{pifont}

\usepackage{subfigure}

\usepackage{multirow}
\usepackage{amsmath}
\usepackage{balance}
\usepackage{amsthm}

\usepackage{fancyhdr,graphicx,amsmath,amssymb}
\usepackage{graphics}

\usepackage{xspace}

\usepackage{colortbl}
\usepackage[table]{xcolor}
\definecolor{grey}{RGB}{2,2,2}
\usepackage{enumitem}

\makeatletter
\def\adl@drawiv#1#2#3{%
        \hskip.5\tabcolsep
        \xleaders#3{#2.5\@tempdimb #1{1}#2.5\@tempdimb}%
                #2\z@ plus1fil minus1fil\relax
        \hskip.5\tabcolsep}
\newcommand{\cdashlineCustom}[1]{%
  \noalign{\vskip\aboverulesep
          \global\let\@dashdrawstore\adl@draw
          \global\let\adl@draw\adl@drawiv}
  \cdashline{#1}
  \noalign{\global\let\adl@draw\@dashdrawstore
          \vskip\belowrulesep}}
\makeatletter

\usepackage{wrapfig}

\usepackage[normalem]{ulem}
\usepackage{hyperref}

\title{Boosting Graph Neural Networks via Adaptive Knowledge Distillation}
\author{
    Zhichun Guo$^{1}$, Chunhui Zhang$^{2}$, Yujie Fan$^{3}$,  Yijun Tian$^{1}$, Chuxu Zhang$^{2}$, Nitesh V. Chawla$^{1}$\\
}
\affiliations{
    $^{1}$University of Notre Dame, Notre Dame, IN 46556\\
    $^{2}$Brandeis University, Waltham, MA 02453\\
    $^{3}$Case Western Reserve University, Cleveland, OH 44106\\
    \{zguo5,yijun.tian,nchawla\}@nd.edu\\ \{chunhuizhang,chuxuzhang\}@brandeis.edu, yxf370@case.edu\\

}

\usepackage{bibentry}

\begin{document}
\include{pythonlisting}

\maketitle

\begin{abstract}

Graph neural networks (GNNs) have shown remarkable performance on diverse graph mining tasks. While sharing the same message passing framework, our study shows that different GNNs learn distinct knowledge from the same graph. This implies potential performance improvement by distilling the complementary knowledge from multiple models. However, knowledge distillation (KD) transfers knowledge from high-capacity teachers to a lightweight student, which deviates from our scenario: GNNs are often shallow. To transfer knowledge effectively, we need to tackle two challenges: how to transfer knowledge from compact teachers to a student with the same capacity; and, how to exploit student GNN's own learning ability. In this paper, we propose a novel adaptive KD framework, called \textit{BGNN}, which sequentially transfers knowledge from multiple GNNs into a student GNN. We also introduce an adaptive temperature module and a weight boosting module. These modules guide the student to the appropriate knowledge for effective learning. Extensive experiments have demonstrated the effectiveness of BGNN. In particular, we achieve up to 3.05\% improvement for node classification and 6.35\% improvement for graph classification over vanilla GNNs.

\end{abstract}

\section{Introduction}

\begin{figure*}[t]
    \centering
    {\includegraphics[ width=0.25\textwidth]{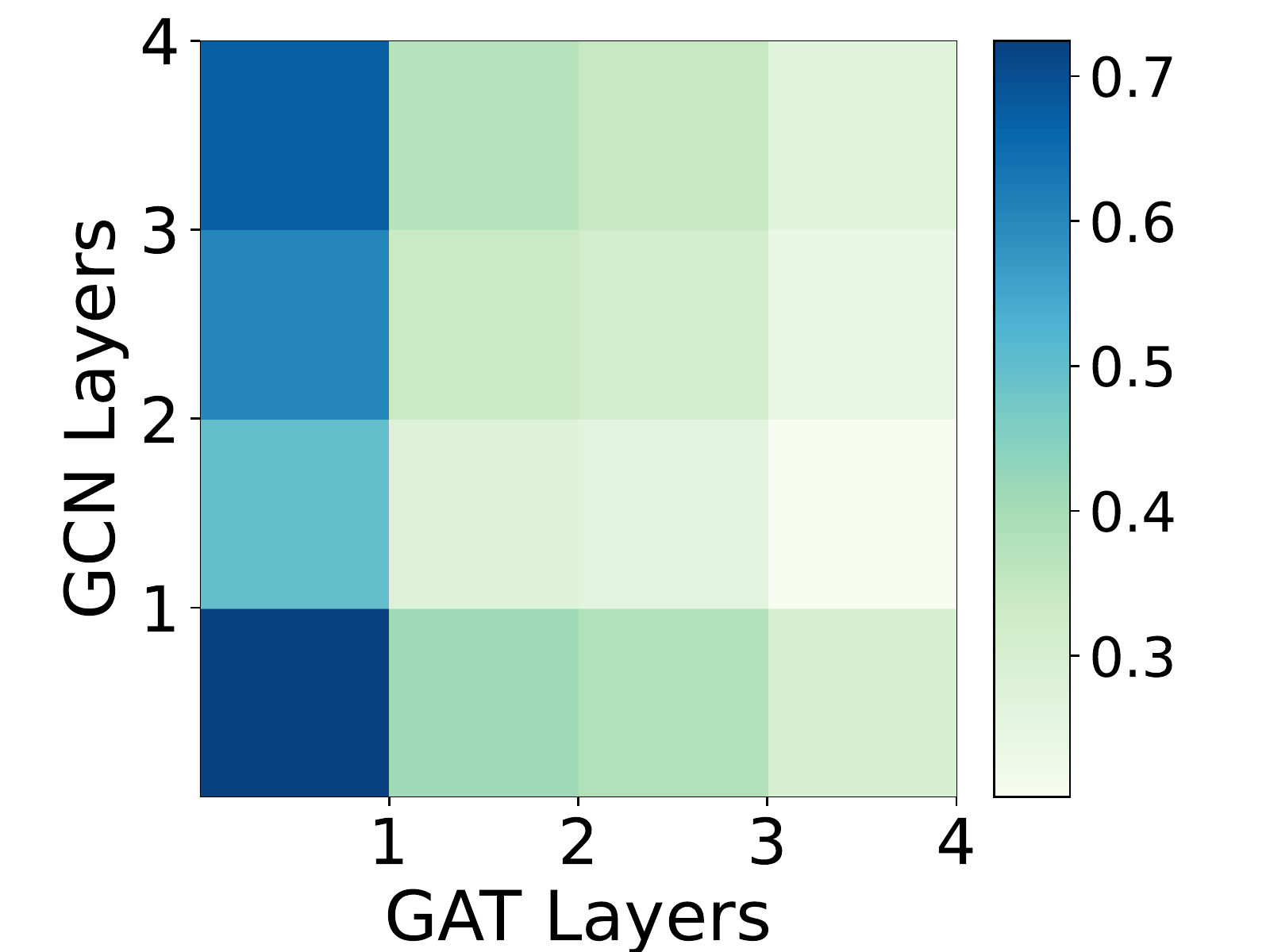}\label{fig:template2}}
    {\includegraphics[ width=0.25\textwidth]{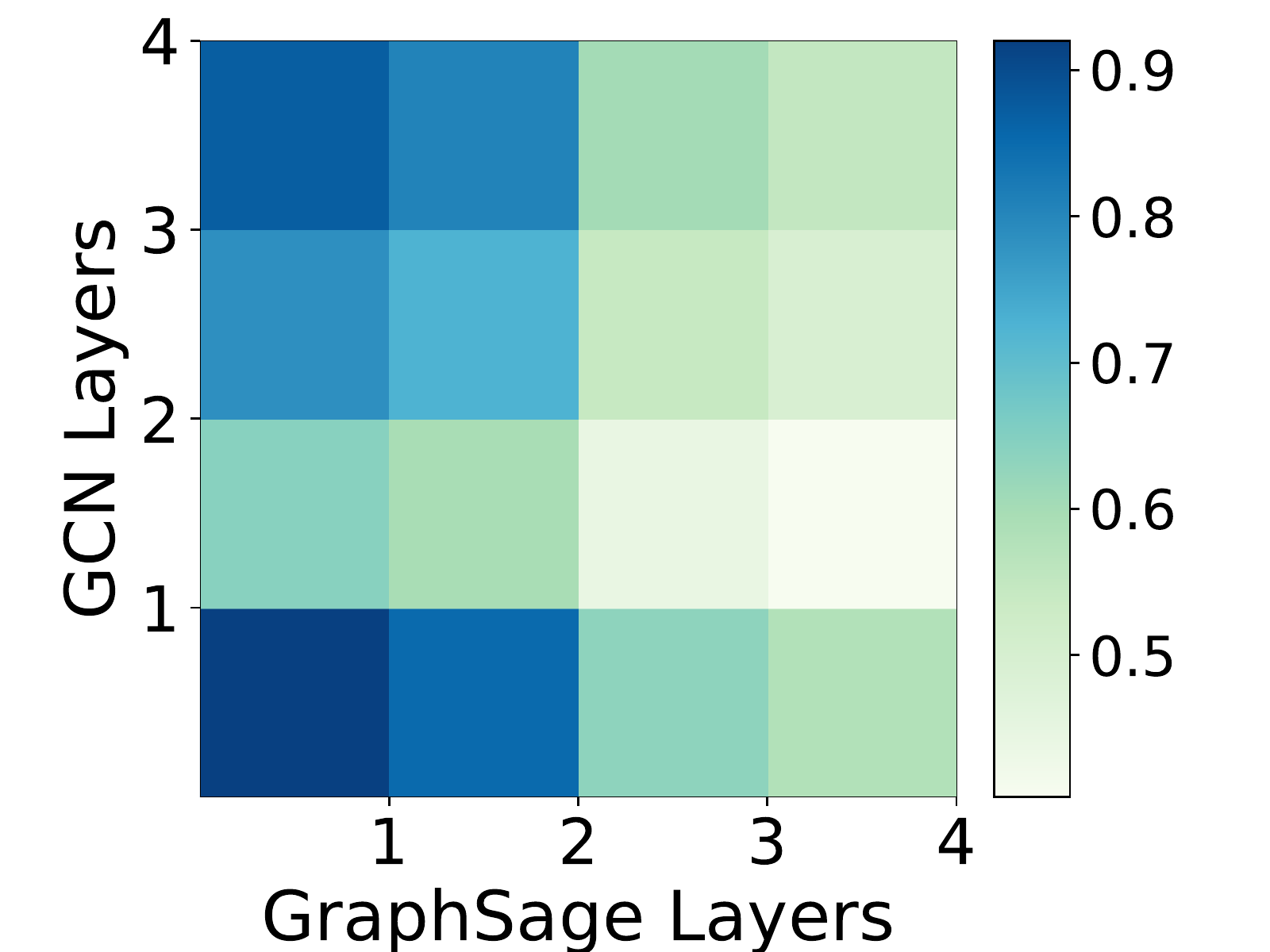}\label{fig:template3}}
    {\includegraphics[width=0.25\textwidth]{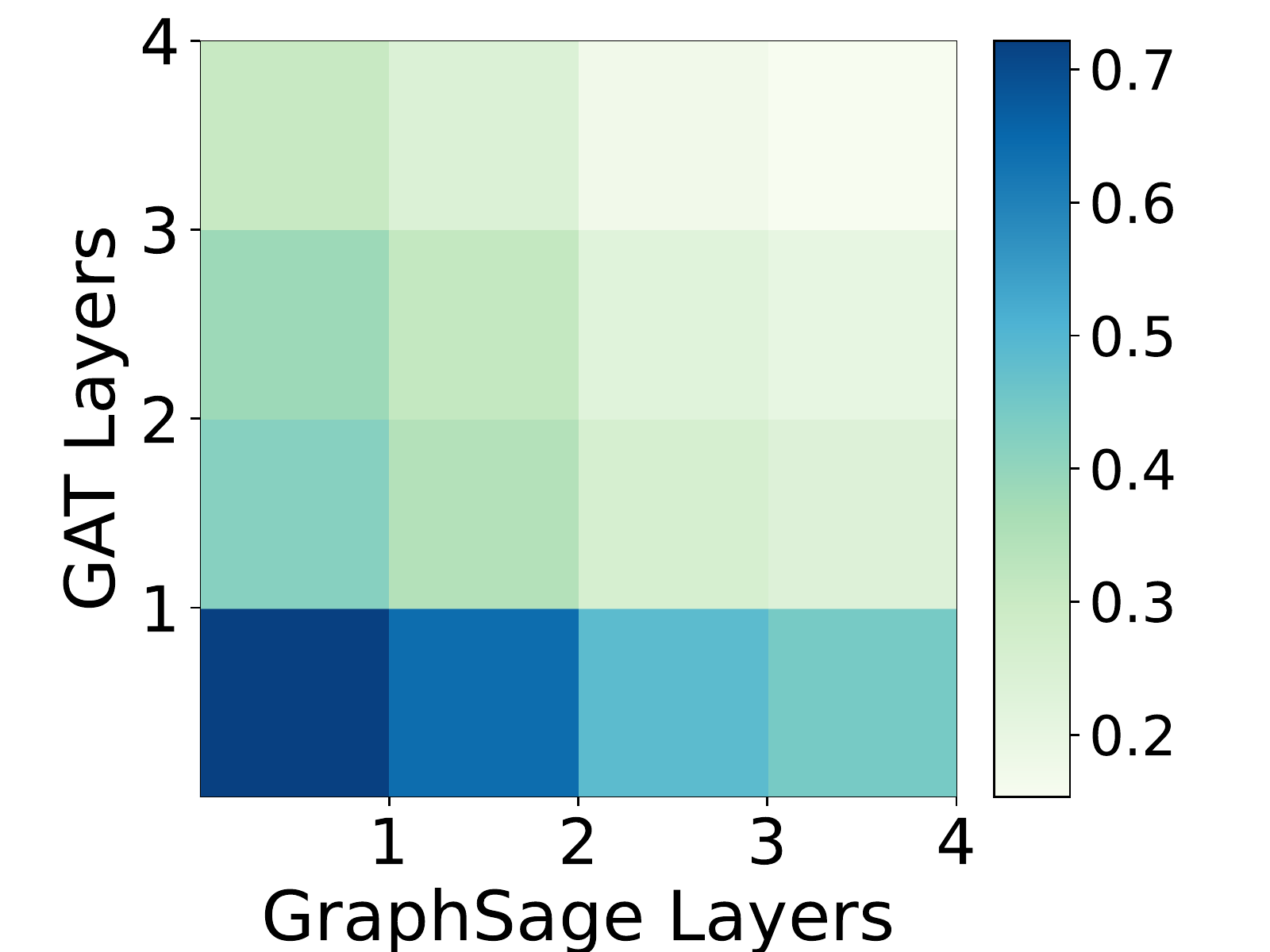}\label{fig:template1}}
    \caption{CKA similarity between graph representation at different layers of GNNs on Enzymes.}
    \label{fig:motiv}
\end{figure*}

Recent years have witnessed the significant development of graph neural networks (GNNs). Various GNNs have been developed and applied to different graph mining tasks~\cite{kipf2017semi, hamilton2017inductive, velickovic2018graph, klicpera2018predict, xu2018powerful, wu2019simplifying,jin2021automated,guo2022graph}. Although most GNNs can be unified into the Message Passing Neural Networks~\cite{gilmer2017neural}, their learning abilities diverge~\cite{xu2018powerful, balcilar2020analyzing}. In our preliminary study, we observe that the graph representations learned by different GNNs are not similar, especially in deeper layers. It suggests that different GNNs may encode complementary knowledge due to their different aggregation schemes. Based on this observation, it is natural to ask: \textit{can we boost vanilla GNNs by effectively utilizing complementary knowledge learned by different GNNs from the same dataset?} 


An intuitive solution is to compose multiple models into an ensemble~\cite{hansen1990neural, breiman2001statistical} that would achieve better performance than each of its constituent models. However, ensemble is not always effective especially when the base classifiers are strong learners~\cite{zhang2020reliable}. Thus, we seek a different approach to take advantage of knowledge from different GNNs: knowledge distillation (KD)~\cite{hinton2015distilling, romero2014fitnets, touvron2021training}, which distills information from one (teacher) model to another (student) model. However, KD is always accompanied by model compression ~\cite{yim2017gift, heo2019comprehensive, yuan2019revisit}, where the teacher network is a high-capacity neural network, and the student network is a compact and fast-to-execute model. Standing by this situation, there could be a significant performance gap between students and teachers. But this kind of performance gap may not exist in our scenario: GNNs are all very shallow due to the oversmoothing issue~\cite{zhao2019pairnorm, li2018deeper, alon2020bottleneck}. Hence, it is more difficult to distill extra knowledge from teacher GNNs to boost the student GNN. To achieve this goal, two major challenges arise: \textit{the first one} is how to transfer knowledge from a teacher GNN into a student GNN with the same capacity that can produce the same even better performance (\textit{teaching effectiveness}); \textit{the second one} is how to push the student model to play the best role in learning by itself, which is ignored in the traditional KD where the student's performance heavily relies on the teacher (\textit{learning ability}).

In this work, we propose a novel framework, namely BGNN, which combines the knowledge from different GNNs in a ``boosting'' way to strengthen a vanilla GNN through knowledge distillation. To improve the teaching effectiveness, we propose two strategies to increase the useful knowledge transferred from the teachers to the student. One is the sequential training strategy, where the student is encouraged to focus on learning from one teacher at a time. This allows the student to learn diverse knowledge from individual GNNs. The other one is an adaptive temperature module. Unlike existing KD methods that use a uniform temperature for all samples, the temperature in BGNN is adjustable based on the teacher's confidence in a specific sample. To enhance the learning ability, we develop a weight boosting module. This module redistributes the weight of samples, making the student GNN pay more attention to the misclassified samples. Our proposed BGNN is a general model which can be applied to both graph classification and node classification tasks. We conduct extensive experimental studies on both tasks, and the results demonstrate the superior performance of BGNN compared with a set of baseline methods.

To summarize, our contributions are listed as follows:

\begin{itemize}[leftmargin=*]
  \item Through empirical study, we show that the representations learned by different GNNs are not similar, indicating that they encode different knowledge from the same input.
  \item Motivated by our observation, we propose a novel framework BGNN that transfers knowledge from different GNNs in a ``boosting'' way to elevate a vanilla GNN.
  \item Rather than using a uniform temperature for all samples, we design an adaptive temperature for each sample, which benefits the knowledge transfer from teacher to student.  
  \item Empirical results have demonstrated the effectiveness of BGNN. Particularly, we achieve up to 3.05\% and 6.35\% improvement over vanilla GNNs for node classification and graph classification, respectively.
\end{itemize}

\section{Related Work and Background}

\textbf{Graph Neural Networks.} Most GNNs follow a message-passing scheme, which consists of message, update, and readout functions to learn node embeddings by iteratively aggregating the information of its neighbors~\cite{xu2018powerful,wu2019simplifying,klicpera2018predict}. For example, GCN~\cite{kipf2017semi} simplifies graph convolutions, which takes averaged aggregation method to aggregate the neighbors' information; GraphSage~\cite{hamilton2017inductive} fixes the number of sampled neighbors to perform aggregation; GAT~\cite{velickovic2018graph} proposes an attention mechanism~\cite{vaswani2017attention} to treat neighbors differently in aggregation. 
The aim of this work is not to design a new GNN architecture, but to propose a new framework to boost existing GNNs by leveraging the diverse learning abilities of different GNNs. 

\noindent\textbf{GNN Knowledge Distillation.} There have been many models that apply KD framework on GNNs for better efficiency in different settings~\cite{yang2021extract, zheng2021cold, zhang2020reliable, denggraph,feng2022freekd}. For example, Yan et al.~\cite{yan2020tinygnn} proposed TinyGNN to distillate a large GNN to a small GNN. GLNN~\cite{zhang2021graph} was proposed to distillate GNNs to MLP. All of these work distillate knowledge by penalizing the softened logit differences between a teacher and a student following~\cite{hinton2015distilling}. Besides this vanilla KD, Yang et al.~\cite{yang2020distilling} proposed LSP, a local structure preserving based KD method in computer vision area, to transfer the knowledge effectively between different GCN models. Wang et al.~\cite{wang2021mulde} propose a novel multi-teacher KD method, MulDE, for link prediction based on knowledge graph embeddings. LLP~\cite{guo2022linkless} is another KD framework specifically for link prediction tasks. In this work, we also take logits-based KD method to distillate knowledge but design two modules to increase useful knowledge transferred from the teachers to the student and play the best of the student model. Different from combining teachers' knowledge in a parallel way in MulDE, we utilize sequential training strategy to combine different teacher models.

\section{Background and Preliminary Study}
\begin{figure*}[t]
    \centering
    {\includegraphics[width=0.8\textwidth]{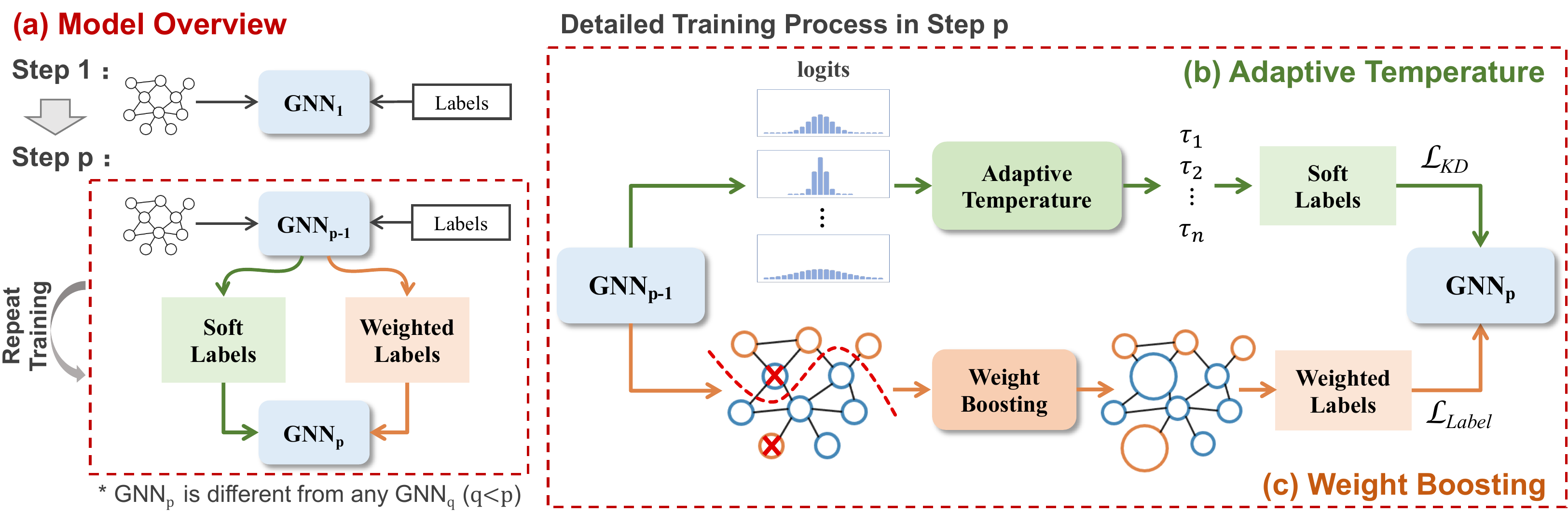}}
    \caption{The overall framework of BGNN. (a) shows our sequential training process, where we train a teacher GNN$_1$ with labels. Then we repeatedly take the generated GNN model from previous step as the teacher to generate soft labels and update the training nodes' weight to train a new initialized student. (b) presents the adaptive temperature module, where the temperature is adjusted based on the teacher's logits distribution for each node. (c) shows the weight boosting module, where the weight of the nodes misclassified by the teacher GNN are boosted (nodes with large size in the above figure).}
   
\label{fig:main}
\end{figure*}

\subsection{Background}
\textbf{Notations.} Let $\mathcal{G} = (\mathcal{V}, \mathcal{E})$ denote a graph, where $\mathcal{V}$ stands for all nodes and $\mathcal{E}$ stands for all edges. Each node $v_i \in \mathcal{V}$ in the graph has a corresponding $D$-dimensional feature vector $\boldsymbol{x}_i \in \mathbb{R}^D$. There are $N$ nodes in the graph. The entire node features matrix is $\textbf{X} \in \mathbb{R} ^ {N \times D}$. 

\noindent\textbf{Graph Neural Networks.} A GNN iteratively updates node embeddings by aggregating information of its neighboring nodes. We initialize the embedding of node $v$ as $\boldsymbol{h}_v^{(0)} = \boldsymbol{x}_v$. Its embedding in the $l$-th layer is updated to $\boldsymbol{h}_v^{(l)}$ by aggregating its neighbors' embedding, which is formulated as: $\boldsymbol{h}^{(l)}_v = \textsc{UPDATE}_l\big(\boldsymbol{h}^{(l-1)}_v,\textsc{AGG}_l\big( \{\boldsymbol{h}^{(l-1)}_u: \forall u \in \mathcal{N}(v) \}\big)\big),$
where $\textsc{AGG}$ and $\textsc{UPDATE}$ are aggregation function and update function, respectively, $\mathcal{N}(v)$ denotes the neighbors of node $v$. Furthermore, the whole graph representation can be computed based on all nodes' representations as: $ \boldsymbol{h}_G = \textsc{READOUT}(\{\boldsymbol{h}^{(l)}_v | v \in \mathcal V\}),$
where $\textsc{readout}$ is a graph-level pooling function.

\noindent\textbf{Node Classification.} Node classification is a typical supervised learning task for GNNs. The target is to predict the label of unlabeled node $v$ in the graph. Let $\textbf{Y} \in \mathbb{R}^{N \times C}$ be the set of node labels. The ground truth of node $v$ will be $y_v$, a $C$-dimension one-hot vector.

\noindent\textbf{Graph Classification.} Graph classification is commonly used in chemistry tasks like molecular property prediction~\cite{hu2019strategies,guo2021few}. Graph classification is to predict the graph properties. Here, the ground truth matrix $\textbf{Y} \in \mathbb{R}^{M \times C}$ is the set of graph labels, where $M$ and $C$ are the number of graphs and graph categories, respectively.


\subsection{Preliminary Study on GNNs' Representation}\label{sec:prelim}
Next, we perform a preliminary study to answer the following question: \textit{do different GNNs encode different knowledge from the same input graphs?} {We train 4-layer GCN, GAT, and GraphSage on Enzymes in a supervised way. After the training, we utilize Centered Kernel Alignment (CKA)~\cite{kornblith2019similarity} as a similarity index to evaluate the relationship among different representations. The higher CKA means the compared representations are more similar. We take the average of all the embeddings at each layer as the representations of that layer. Figure~\ref{fig:motiv} illustrates the CKA between representations of each layer learned from GCN, GAT, and GraphSage on Enzymes. We observe that the similarities between the learned representations at different layers in GCN, GAT, and GraphSage are diverse. } For example, the CKA value between the representation from layer 1/2/3/4 of GCN and that from GAT is around 0.7/0.35/0.4/0.3. It indicates that different GNNs may encode different knowledge.

We posit that different aggregation schemes in these GNNs cause difference in the learned representations. In particular, GCN aggregates neighborhoods with predefined weights; GAT aggregates neighborhoods using learnable weights and GraphSage randomly samples neighbors during aggregation. Given such differences, it is promising to boost one GNN by incorporating the knowledge from other GNNs and it motivates us to design a framework that can take advantage of the diverse knowledge from different GNNs. 

\section{The Proposed Framework}

In this section, we introduce the framework BGNN to boost GNNs by utilizing complementary knowledge from other GNNs. An illustration of the model framework is shown in Figure~\ref{fig:main}. Our framework adopts a sequential training strategy to encourage the student to focus on learning from one single teacher at a time. To adjust the information distilled from the teacher, we propose an adaptive temperature module to adjust the soft labels from teachers. Further, we propose a weight boosting mechanism to enhance the student model training.

\subsection{Model Overview}

To boost a GNN, we take it as the student and we aim to transfer diverse knowledge from other teacher GNNs into it. In this work, we utilize the KD method proposed by~\cite{hinton2015distilling}, where a teacher's knowledge is transferred to a student by encouraging the student model to imitate the teacher's behavior. In our framework, we pre-train a teacher GNN (GNN$_\mathrm{T}$) with ground-truth labels and keep its parameters fixed during KD. Then, we transfer the knowledge from GNN$_\mathrm{T}$ by letting the student GNN (GNN$_\mathrm{S}$) optimize the soft cross-entropy loss between the student network's logits $\pmb{z}_v$ and the teachers' logits $\pmb{t}_v$. 
Let $\tau_v$ be the temperature for node $v$ to soften the logits distribution of teacher GNN. 

Then we incorporate the target of knowledge distillation into the training process of the student model by minimizing the following loss:
\begin{equation}
\begin{aligned}
    \mathcal{L} = \mathcal{L}_\text{label} + \lambda
    \mathcal{L}_{KD} =  \mathcal{L}_\text{label} - \lambda \sum\nolimits_{v \in \mathcal{V}}\hat{\pmb{y}}_v^\mathrm{T}\mathrm{log}(\hat{\pmb{y}}_v^\mathrm{S}), \\
\text{with} \quad \hat{\pmb{y}}_v^\mathrm{T} = \mathrm{softmax}(\pmb{t}_v/\tau_v), \; \hat{\pmb{y}}_v^\mathrm{S} = \mathrm{softmax}(\pmb{z}_v/\tau_v),
\end{aligned}
\end{equation}
where $\mathcal{L}_\text{label}$ is the supervised training loss w.r.t. ground-truth labels and $\lambda$ is a trade-off factor to balance their importance. 

Following the above objective, we adopt a sequential process of distillation, where we can freely boost the student GNN with one teacher GNN or multiple teachers. Such a sequential manner encourages the student model to focus on the knowledge from one single teacher. In contrast, when using multiple teachers simultaneously, the student may receive mixed noisy signals which can harm the distillation process.


As illustrated in Figure~\ref{fig:main}(a), we first train a teacher GNN$_1$ using true labels and train a student GNN$_2$ with dual targets of predicting the true labels and matching the logits distribution of GNN$_1$. The logits distribution has been softened by our proposed adaptive temperature (Figure~\ref{fig:main}(b)) for each node and the weight of the nodes misclassified by the teacher GNN are boosted when predicting true labels (Figure~\ref{fig:main}(c)). The parameters of teacher GNN$_1$ are not updated when we train GNN$_2$. Such process is repeated for $p$ steps and the knowledge from GNN$_{p-1}$, $\ldots$, GNN$_{1}$ can be transferred into the last student GNN$_{p}$.

\subsection{Distillation via Adaptive Temeperature}
\label{sec: temp}

Instead of leveraging KD for compressing GNN models, we aim to use KD for boosting the student by transferring knowledge between GNNs sharing the same capacity. To achieve this goal, we need a more powerful KD method to fully take advantage of the useful knowledge in the teacher GNN so as to produce better performance. 

\noindent\textbf{Analysis of KD.} Before we introduce the detailed technique, we first analyze the reason behind KD's success by comparing the gradient of $\mathcal{L}_{KD}$ and supervised loss. Specifically, for any single sample, we compute the gradient of $\mathcal{L}_{KD}$ for its $c$-th output (Detailed derivation is in Section A of Appendix): 
\begin{align}
\frac{\partial{\mathcal{L}_{KD}}}{\partial{z_{v,c}}} &= -\frac{\partial}{\partial z_{v,c}} \sum\nolimits_{j=1}^{C} \hat{y}_{v,j}^\mathrm{T}\mathrm{log}(\hat{y}_{v,j}^\mathrm{S}) = \frac{1}{\tau_v}(\hat{y}_{v,c}^\mathrm{S} - \hat{y}_{v,c}^\mathrm{T}) \nonumber \\
    &=\frac{1}{\tau_v}\Big(\frac{e^{z_{v,c/\tau_v}}}{\sum\nolimits_{j=1}^{C}e^{z_{v,j}/\tau_v}} - \frac{e^{t_{v,c/\tau_v}}}{\sum_{j=1}^{C}e^{t_{v,j}/\tau_v}}\Big).
    \label{equation:anal}
\end{align}
Let $*$ denote the true label for the single sample, i.e., $y_{v,*} = 1$. Then the gradient of KD loss for this sample is:
\begin{align}
\frac{\partial{\mathcal{L}_{KD}}}{\partial{z_v}} &= \frac{1}{\tau_v}\sum\nolimits_{c=1}^{C}(\hat{y}_{v,c}^\mathrm{S} - \hat{y}_{v,c}^\mathrm{T}) \nonumber \\
    =\frac{1}{\tau_v} & \Big((\hat{y}_{v,*}^\mathrm{S} - \hat{y}_{v,*}^\mathrm{T}) + \sum\nolimits_{c=1, c \neq *}^{C} (\hat{y}_{v,c}^\mathrm{S} - \hat{y}_{v,c}^\mathrm{T})\Big).
\end{align}
In the above equation, the first term $(\hat{y}_{v,*}^\mathrm{S} - \hat{y}_{v,*}^\mathrm{T})$ corresponds to transferring the knowledge hidden in the distribution of the logits of the correct category; the second term is responsible for transferring the dark knowledge from the wrong categories. This dark knowledge contains important information about the similarity between categories, which is the key point for the success of KD~\cite{hinton2015distilling}. We rewrite the first term as: $\hat{y}_{v,*}^\mathrm{S} - \hat{y}_{v,*}^\mathrm{T}y_{v,*}$. Note that the gradient for the sample of the cross entropy between student logits $z_v$ and ground truth label is $\hat{y}_{v,*}^\mathrm{S} - y_{v,*}$ when $\tau_v$ is 1. We can view $\hat{y}_{v,*}^\mathrm{T}$ as the importance weight of the true label information. If the teacher is confident for this sample (i.e. $\hat{y}_{v,*}^\mathrm{T} \approx 1$), the ground truth related information will play a more important role in gradient computation. Thus, both true label information and similarity information hidden in the wrong categories contribute a lot to the success of KD. The confidence of the teacher is the key to balancing these two parts' importance in KD training process.

\noindent\textbf{Detailed Technique.} From the above analysis, temperature directly controls the trade-off between true label knowledge and dark knowledge. Additionally, existing work~\cite{zhang2020self} has proved that temperature scaling helps to yield more calibrated models. To transfer more useful knowledge from teachers to students, we make the predefined hyper-parameter temperature adaptive for all nodes:  each node is associated with an adaptive temperature based on the confidence of teachers for each node. Specifically, we take the entropy of the teacher's logits $\pmb{t}_v$ to measure the teachers' confidence for each node. The lower entropy means more confident the teacher is for the specific node. Then we compute the temperature for each node with learnable parameters by the following formulation:
\begin{align}
   & \mathrm{Confidence}(\pmb{t}_v) = -\sum\nolimits_{c=1}^{C}t_{v,c}log(t_{v,c}), \\
   & \tau_v = \sigma\Big(\mathrm{MLP}\big(\mathrm{Confidence}(\pmb{t}_v)\big)\Big),
\end{align}
where $\sigma$ represents $\mathrm{sigmoid}$ operation. We use $\tau_v = \tau_v \times (\tau_{max} - \tau_{min}) + \tau_{min}$ to limit the temperature into a fixed range [$\tau_{min}, \tau_{max}$]. In the experiments, we discover that involving the distribution of the logits of the teacher in the temperature is more effective than only considering the teachers' confidence after two training steps:
\begin{equation}
    \tau_v = \sigma\Big(\mathrm{MLP}\big(\textsc{Concat}(\pmb{t}_v, \mathrm{Confidence}(\pmb{t}_v))\big)\Big),
\end{equation}
where $\textsc{Concat}$ represents the concatenation function.

\subsection{Weight Boosting}

As we mentioned earlier, we aim to transfer knowledge from teachers to students with the same capacity. In this case, we should not only enhance the knowledge transferred from teachers to the student but also enhance the supervised training process of the student GNN itself.
For the supervised training procedure, inspired by Adaboost algorithm~\cite{freund1999short, sun2019adagcn}, we propose to boost the weights of misclassified nodes by the teacher GNN, which encourages the student GNN to pay more attention to these misclassified samples and learn them better. For better efficiency, we drop the ensemble step of boosting and take the student GNN generated by the last training step for the latter prediction in test data. In specific, firstly, we initialize the node weights $w_i = 1/N_{train}, i = 1,2,3,...,N_{train}$ on training set. Next, we pre-train a teacher GNN with samples of the training set in a supervised way. Then we boost the weights of training samples (i.e. graphs for graph classification and nodes for node classification) that are misclassified by GNN$_\mathrm{T}$. Here, we apply SAMME.R~\cite{hastie2009multi} to update the weight:
\begin{equation}
    w_i = w_i \cdot \mathrm{exp}\Big(-\frac{C-1}{C}\pmb{y}_i^\top \mathrm{log}\ \pmb{p}_i\Big), i = 1,2,..., N_{train},
\end{equation}
where $\pmb{p}_i = \mathrm{softmax}(\pmb{z}_i)$. 

\vskip 0.5em
\noindent\textbf{Overall Objective.}
By incorporating $w_i$ into the supervised training process and combining KD training, the loss function of our method can be formulated as:
\begin{align}
     &\mathcal{L}_{BGNN} = \mathcal{L}_{Label} + \lambda \mathcal{L}_{KD} \nonumber \\
     &= - \sum\nolimits_{i=1}^{N_{train}}w_i\ \pmb{y}_i\mathrm{log}(\pmb{p}_i) - \lambda \sum_{v \in \mathcal{V}}\hat{\pmb{y}}_v^\mathrm{T}\mathrm{log}(\hat{\pmb{y}}_v^\mathrm{S}).
\end{align}

\section{Experiments}

\begin{table*}[t]
    \centering 
    \scalebox{0.9}{
    \begin{tabular}{>{\small}c|>{\small}c|>{\small}c>{\small}c>{\small}c|>{\small}c>{\small}c>{\small}c>{\small}c}
    \toprule
    \multicolumn{2}{c|}{}&\multicolumn{3}{c|}{Graph Classification}&\multicolumn{4}{c}{Node Classification}\\
    \midrule
    Student & Method & \textsc{Collab} & \textsc{IMDB} & \textsc{Enzymes}  & \textsc{Cora} & \textsc{Citeseer} & \textsc{Pubmed} & \textsc{A-Computers}\\
    \midrule
    {\multirow{4}{*}{GCN}} & NoKD & 80.82$\pm$0.99 & 77.20$\pm$0.40 & 66.33$\pm$1.94 & 82.25$\pm$0.39 & 72.30$\pm$0.21 & 79.40$\pm{0.56}$ & 87.79$\pm$0.04\\
    &KD~\cite{hinton2015distilling} & 81.24$\pm$0.63 & 77.60$\pm$0.77 & 65.00$\pm$1.36  & 82.78$\pm$0.42 & 72.59$\pm$0.28  & 79.60$\pm$0.67 & 87.13$\pm$0.11\\
    &LSP~\cite{yang2020distilling} & 81.22$\pm$0.29 & 77.99$\pm$1.21 & 67.12$\pm$1.11 & 82.29$\pm$0.77 & 72.77$\pm$0.34 & 78.93$\pm$0.38 & 87.93$\pm$0.72\\
    \cmidrule{2-9}
    &BGNN & \cellcolor{grey!10}\textbf{82.73$\pm$0.34} & \textbf{79.33$\pm$0.47} & \textbf{69.44$\pm$0.79} & \textbf{83.97$\pm$0.17} & \textbf{73.87$\pm$0.24} & \cellcolor{grey!10}\textbf{80.73$\pm$0.25} & \textbf{89.58$\pm$0.03}\\
    \midrule
    \midrule
    {\multirow{4}{*}{SAGE}} & NoKD & 81.20$\pm$0.55 & 77.60$\pm$1.02 & 73.67$\pm$5.52 & 81.18$\pm$0.92 & 71.62$\pm$0.33 & 78.00$\pm$0.23 & 87.49$\pm$1.40 \\ 
    &KD~\cite{hinton2015distilling} & 80.34$\pm$0.50 & 77.90$\pm$1.14 & 71.67$\pm$1.36 & 81.72$\pm$1.04 & 72.58$\pm$0.32 & 77.25$\pm$0.39 & 89.34$\pm$0.18\\
    &LSP~\cite{yang2020distilling} & 81.33$\pm$0.31 & 78.23$\pm$0.91 & 74.23$\pm$2.16 & 82.00$\pm$0.75 & 71.37$\pm$0.34 & 77.40$\pm$0.22 & 88.49$\pm$0.72\\
    \cmidrule{2-9}
    &BGNN & \textbf{82.67$\pm$0.57} & \cellcolor{grey!10}\textbf{79.67$\pm$0.47} & \textbf{78.33$\pm$1.00} & \textbf{83.30$\pm$0.35} & \textbf{73.90$\pm$0.16} & \textbf{79.03$\pm$0.09} & \cellcolor{grey!10}\textbf{89.85$\pm$0.12}\\
    \midrule
    \midrule
    {\multirow{4}{*}{GAT}} & NoKD & 79.28$\pm$0.47 & 76.20$\pm$1.33 & 74.33$\pm$2.00 &83.44$\pm$0.17 & 72.46$\pm$0.75 & 79.36$\pm$0.19 & 87.98$\pm$0.29 \\ 
    &KD~\cite{hinton2015distilling} & 78.98$\pm$1.01 & 77.80$\pm$1.25 & 72.67$\pm$2.36 & 83.50$\pm$0.27 & 72.52$\pm$0.20 & 78.91$\pm$0.18 & 86.90$\pm$0.25 \\
    &LSP~\cite{yang2020distilling} &78.99$\pm$1.39 &78.00$\pm$0.93 & 74.43$\pm$3.27& 83.36$\pm$0.57 & 71.90$\pm$0.36 &79.53$\pm$0.26 &86.90$\pm$0.48\\
    \cmidrule{2-9}
    &BGNN & \textbf{80.73$\pm$0.25} & \textbf{79.33$\pm$0.94} & \cellcolor{grey!10}\textbf{79.36$\pm$2.81} & \cellcolor{grey!10}\textbf{84.63$\pm$0.28} & \cellcolor{grey!10}\textbf{74.53$\pm$0.29} & \textbf{79.83$\pm$0.12} & \textbf{89.43$\pm$0.11}\\
    \bottomrule
    \end{tabular}
    }
    \caption{Classification performances under the single teacher setting. NoKD denotes the results of supervised training without KD. Bold text is used to highlight best results for each backbone. Shadow is to highlight best results for each dataset.}
    \label{tab:baseline_single}
\end{table*}

\begin{table*}[!t]
    \centering
    \scalebox{0.87}{
    \begin{tabular}{>{\small}c|>{\small}c|>{\small}c>{\small}c>{\small}c|>{\small}c>{\small}c>{\small}c>{\small}c}
    \toprule
    \multicolumn{2}{c|}{}&\multicolumn{3}{c|}{Graph Classification}&\multicolumn{4}{c}{Node Classification}\\
    \midrule
    Student & Method & \textsc{Collab} & \textsc{IMDB} & \textsc{Enzymes}  & \textsc{Cora} & \textsc{Citeseer} & \textsc{Pubmed} & \textsc{A-Computers}\\
    \midrule
    \midrule
    {\multirow{5}{*}{GCN}} &BAN~\cite{furlanello2018born} & 81.60$\pm$0.40 & 78.50$\pm$1.00 & 66.67$\pm$1.67 & 83.17$\pm$0.26 & 72.47$\pm$0.21& 79.87$\pm$0.21 & 88.78$\pm$0.05\\
    &MulDE~\cite{wang2021mulde} &80.86$\pm$1.18 & 77.50$\pm$1.20 & 67.33$\pm$0.90 & 82.53$\pm$0.05 &72.33$\pm$0.09 & 78.73$\pm$0.09 & 88.41$\pm$0.12\\
    \cmidrule{2-9}
    &BGNN(s) & 82.73$\pm$0.34 & 79.33$\pm$0.47 & 69.44$\pm$0.79 & 83.97$\pm$0.17 & 73.87$\pm$0.24 & 80.73$\pm$0.25 & \textbf{89.58$\pm$0.03} \\
    &BGNN(m)-ST & 82.87$\pm$0.09 & \textbf{79.67$\pm$0.27} & \textbf{71.12$\pm$2.45} & \cellcolor{grey!10}\textbf{84.83$\pm$0.25} & 73.60$\pm$0.14 & 80.20$\pm$0.08 & 89.03$\pm$0.02\\
    &BGNN(m)-TS & \cellcolor{grey!10}\textbf{83.40$\pm$0.15} & 79.00$\pm$0.00 & 70.00$\pm$1.36 & 84.40$\pm$0.22 & \cellcolor{grey!10}\textbf{74.87$\pm$0.25} & \cellcolor{grey!10}\textbf{80.90$\pm$0.00} & 89.02$\pm$0.03\\
    \midrule
    \midrule
    {\multirow{5}{*}{SAGE}} &BAN~\cite{furlanello2018born} & 81.80$\pm$0.20 & \textbf{80.73$\pm$0.20} & 70.00$\pm$1.67 & 82.80$\pm$0.31 & 73.10$\pm$0.57 & 77.90$\pm$0.08 & 89.73$\pm$0.08\\
    & MulDE~\cite{wang2021mulde} & 81.00$\pm$0.91 & 77.60$\pm$1.50 & 77.00$\pm$0.14 & 81.67$\pm$0.09 & 68.83$\pm$0.34 & 78.13$\pm$0.34 & 88.05$\pm$1.35\\ 
    \cmidrule{2-9}
    &BGNN(s) & 82.67$\pm$0.57 & 79.67$\pm$0.47 & \textbf{78.33$\pm$1.00} & 83.30$\pm$0.35 & 73.90$\pm$0.16 & 79.03$\pm$0.09 & 89.85$\pm$0.12 \\
    &BGNN(m)-TC &82.80$\pm$0.28 & 79.67$\pm$0.49 & 78.03$\pm$0.49 & \textbf{83.90$\pm$0.22} & \textbf{74.67$\pm$0.33} & \textbf{79.23$\pm$0.05} & 90.12$\pm$0.05\\
    &BGNN(m)-CT &\textbf{83.30$\pm$0.23} & 79.33$\pm$0.79 & 77.78$\pm$1.57 & 83.90$\pm$0.00 &74.40$\pm$0.16 & 79.10$\pm$0.08 & \cellcolor{grey!10}\textbf{90.40$\pm$0.16}\\
    \midrule
    \midrule
    {\multirow{5}{*}{GAT}} &BAN~\cite{furlanello2018born} & 80.10$\pm$0.50& 79.50$\pm$0.50 & 76.83$\pm$1.50 & 83.93$\pm$0.21 & 72.90$\pm$0.08 & 79.57$\pm$0.12 & 87.66$\pm$0.15\\
    & MulDE~\cite{wang2021mulde} & 79.40$\pm$1.65 & 78.50$\pm$0.43 & 77.50$\pm$0.58 & 83.23$\pm$0.33 & 71.23$\pm$0.19 & 78.40$\pm$0.08 & 88.08$\pm$0.18\\ 
    \cmidrule{2-9}
    &BGNN(s) & 80.73$\pm$0.25 & 79.33$\pm$0.94 & 79.36$\pm$2.81 & 84.63$\pm$0.28 & 74.53$\pm$0.29 & 79.83$\pm$0.12 & \textbf{89.43$\pm$0.11} \\
    &BGNN(m)-SC &81.53$\pm$0.41 & \cellcolor{grey!10}\textbf{81.33$\pm$1.79} & 78.89$\pm$0.79 & \textbf{84.77$\pm$0.05} & 74.20$\pm$0.08 & \textbf{80.30$\pm$0.17} & 89.27$\pm$0.04\\
    &BGNN(m)-CS & \textbf{81.80$\pm$0.09} & 80.33$\pm$0.94 & \cellcolor{grey!10}\textbf{80.68$\pm$1.49} & 84.30$\pm$0.22 & \textbf{74.70$\pm$0.08} & 79.83$\pm$0.05 & 89.07$\pm$0.11\\
    \midrule
    \midrule
    \multicolumn{2}{c|}{Ensemble~\cite{hansen1990neural}} & 82.27$\pm$0.09 & 78.67$\pm$0.94 & 80.00$\pm$3.60 & 82.63$\pm$0.05 & 72.43$\pm$0.26 & 78.17$\pm$0.31 & 90.40$\pm$0.30\\
    \bottomrule
    \end{tabular}
    }
    \caption{Classification performances under the multi teacher setting. For simplicity, we abbreviate GCN to C, GraphSage to S, and GAT to T. The order in which the abbreviation letters appear reflects their training order. For example, BGNN-ST indicates that GraphSage (S) and GAT (T) are the teachers in turn. We further denote the GNNs trained with single teacher as BGNN(s) and the GNNs trained with multiple teachers as BGNN(m). Best performances for each backbone and each dataset are marked with bold and shadow, respectively.}
    \label{tab:baseline_multi}
\end{table*}



\subsection{Experimental Setup}
\newcommand{\specialcell}[2][c]{\begin{tabular}[#1]{@{}c@{}}#2\end{tabular}}


\textbf{Datasets.} We use seven datasets to conduct graph classification and node classification experiments. We follow the data split as in original papers~\cite{sen2008collective, namata2012query} for Cora, Citeseer and Pubmed while the remaining datasets are randomly split using an empirical ratio. The more details are in Section B of Appendix.

\noindent\textbf{BGNN Training.} We select GraphSage~\cite{hamilton2017inductive}, GCN~\cite{kipf2017semi} and GAT~\cite{velickovic2018graph} as GNN backbones and examine the performance of BGNN with different combinations of teachers and students. For all the GNN backbones, we use two-layer models. For the single teacher setting, we use one GNN as the teacher and a different GNN as the student. For the multiple teacher setting, we permute the three GNNs in six orders, where the first two serve as teachers and the third one is the student. During the training, we clamp the adaptive temperature in the range from 1 to 4, which is a commonly used temperature range in KD. Implementation and experiment details are shown in Section C of Appendix.


\noindent\textbf{Baselines.} For the single teacher setting, we compare BGNN with the student GNN trained without a teacher and the student GNN trained with different teacher GNNs using KD~\cite{hinton2015distilling} and LSP~\cite{yang2020distilling}. {KD is the most commonly used framework to distillate the knowledge from teacher GNNs to student GNNs. LSP is the state-of-the-art KD framwork proposed for GNNs, which teaches the student based on local structure information instead of logits information.} We set the temperature $\tau$ in KD to a commonly used value 4~\cite{stanton2021does}. For the multi teacher setting, we compare BGNN with two multi-teacher KD frameworks (i.e., BAN~\cite{furlanello2018born} and MulDE~\cite{wang2021mulde}). BAN~\cite{furlanello2018born} uses a similar sequential training strategy as ours. It follows the born-again idea~\cite{breiman1996born} to train one teacher and multiple students sharing the same architecture, and the resulting model is an ensemble~\cite{hansen1990neural} of all trained students. MulDE~\cite{wang2021mulde} trains the student by weighted-combining 
multiple teachers' outputs and utilizing KD. Here, the teachers' outputs are their generated logits for each node or graph. Additionally, we also include the ensemble that averages the prediction of supervised GCN, GraphSage and GAT as a baseline.

\noindent\textbf{Evaluation.} We evaluate the performance by classification accuracy. For graph classification tasks, we obtain the entire graph representation by sum pooling as recommended by~\cite{xu2018powerful}.
For each experiment, we report the average and standard deviation of the accuracy from ten training rounds. 

\subsection{How Does BGNN Boost the Vanilla GNNs?}
This experiment examines whether our BGNN can boost the performance of the vanilla GNNs by transferring knowledge from multiple GNNs. Both the single teacher setting and the multiple teacher setting are considered.

\noindent\textbf{For the single teacher setting,} there are six possible teacher-student pairs: GCN-GAT, GCN-GraphSage, GraphSage-GCN, GraphSage-GAT, GAT-GCN and GAT-GraphSage. Each teacher-student pair is trained using our BGNN, KD and LSP. Note that for each student GNN, two teacher options are available. We report the better result of the two teachers for each of BGNN, KD and LSP, which is different from the setting of our ablation study. In Table~\ref{tab:baseline_single}, we can observe that BGNN outperforms the respective supervisedly trained GNN (NoKD) by significant margins for all datasets, regardless of the selection of student GNN. This confirms that our BGNN can boost the vanilla GNNs by transferring additional knowledge from other GNNs. However, we should also note that, the prediction power of the student GNN is still constrained by its own architecture. For example, for the Enzymes dataset, the original GAT without distillation (74.33\%) outperforms the original GCN (66.33\%) by a large margin. Although our BGNN can boost the performance of GCN from 66.33\% to 69.44\%, the boosted accuracy of GCN is still lower than that of the original GAT. This indicates that selecting an appropriate student GNN is still important. 

\begin{figure}[t]
    \centering
    \includegraphics[width=0.45\textwidth]{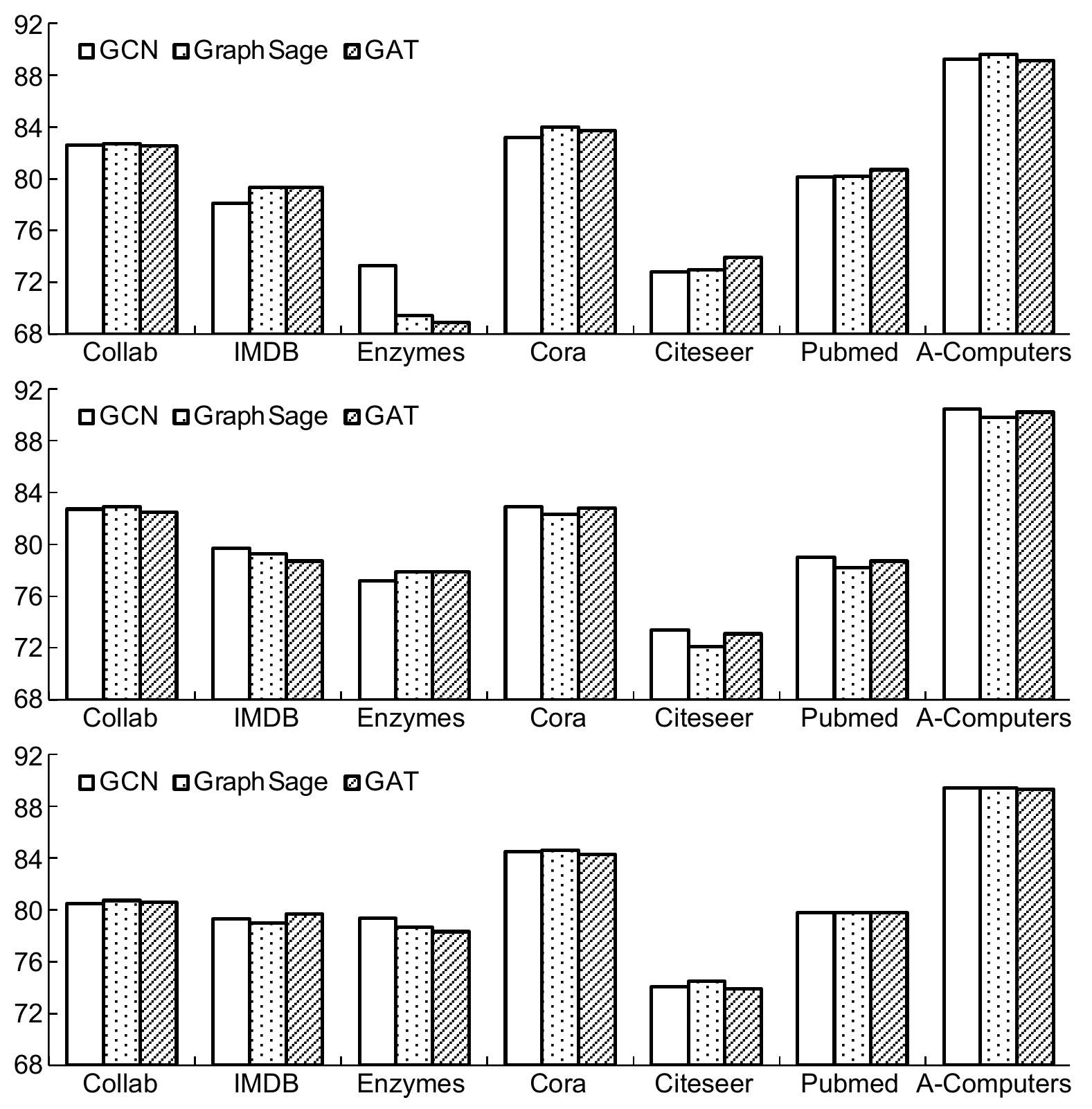}
    \caption{Results of BGNN with different teacher and student GNNs pairs. \textit{Upper}: GCN as student. \textit{Middle}: GraphSage as student. \textit{Lower}: GAT as student.}
    \label{fig:teach_student}
\end{figure}

\noindent\textbf{For the multi-teacher setting,} we examine the performance of BGNN with all the six permutations of the three GNNs. In Table~\ref{tab:baseline_multi}, we find that BGNN(m) outperforms the respective BGNN(s) in most cases, and the supervised trained GNNs (NoKD) in all cases. This indicates that the extra teachers in the BGNN training successfully transfer additional knowledge into the student GNNs. 

We further compare BGNN(m) with the average ensemble of the vanilla GNNs, which combines the knowledge of GNNs in a straightforward manner. In Table~\ref{tab:baseline_multi}, BGNN(m) obtains better prediction results than those of the ensemble. The ensemble fails to leverage the mutual knowledge in the training stage, as it combines the prediction results directly. In contrast, our BGNN trains the student GNNs under the guidance of teachers. This combines the knowledge from teachers and students in a natural way, providing better prediction power to the student trained with BGNN.

\begin{table*}[t]
\centering
{\scalebox{1.0}{
\begin{tabular}{>{\small}c>{\small}c>{\small}c>{\small}c>{\small}c>{\small}c>{\small}c>{\small}c}
\toprule
 Variant & \textsc{Collab} & \textsc{IMDB} & \textsc{Enzymes}  & \textsc{Cora} & \textsc{Citeseer} & \textsc{Pubmed} & \textsc{A-Computers}\\
\midrule
w/o Adaptive Temp & 81.27$\pm$0.25 & 79.00$\pm$0.00 & 66.67$\pm$1.36 & 83.30$\pm$0.22 & 73.18$\pm$0.23 & 80.10$\pm$0.14 & 87.30$\pm$0.06 \\
w/o Weight Boosting & 82.53$\pm$0.09 & 77.67$\pm$0.47 & 68.33$\pm$1.36 & 83.17$\pm$0.42 & 72.80$\pm$0.17 & 79.80$\pm$0.14 & 88.56$\pm$0.00\\
BGNN & \textbf{82.53$\pm$0.47} & \textbf{79.33$\pm$0.47} & \textbf{68.89$\pm$0.79} & \textbf{83.70$\pm$0.37} & \textbf{73.87$\pm$0.24} & \textbf{80.73$\pm$0.25} & \textbf{89.12$\pm$0.01}\\
\bottomrule
\end{tabular}}}
\caption{Results of fusion selections (BGNN, BGNN without adaptive temperature, and BGNN without weight boosting). Here, we use GAT as the teacher and GCN as the student, the same as for Figure~\ref{fig:ablation_temp} and Figure~\ref{fig:ablation_boosting}.}
\label{tab:ablation}
\end{table*}

\subsection{Does BGNN Distill Knowledge More Effectively?} 
\textbf{For the single teacher setting,} we compare our BGNN with KD~\cite{hinton2015distilling} and LSP~\cite{yang2020distilling}. In Table~\ref{tab:baseline_single}, we observe that our method can outperform both KD and LSP on both graph classification and node classification tasks. This may credit to both the adaptive temperature and weight boosting modules. 

\noindent\textbf{For the multi-teacher setting,} we compare BGNN with MulDE and BAN. MulDE combines the teacher models in parallel, while our BGNN and BAN combines the teacher models in a sequential manner. As shown in  Table~\ref{tab:baseline_multi}, our BGNN achieves the best performance for all the seven datasets. BGNN also performs better than the respective MulDE and BAN in almost all settings (20 out of 21 cases). In addition, we find that the sequential methods (BGNN and BAN) outperform the parallel method (MulDE) in general. 

A possible explanation is that all GNNs play an important role in the parallel MulDE training at the same time, which may not exploit the full potential of the more powerful GNNs. On the contrary, the sequential methods trains a student with a single teacher at each step. This allows the student to focus on learning from one specific teacher GNN. In this way, the prediction power of a single powerful teacher GNN may be transferred to the student more effectively. 

For the sequential methods, BAN uses the same architecture for both the teachers and the students. Therefore, it is not able to leverage information from different GNNs. In contrast, our BGNN achieves better results by combining knowledge of different models. Furthermore, it is worth mentioning that our model does not use the average ensemble step like BAN. This may lead to higher accuracy as well, assuming that the final student can inherit the power of all preceding GNNs.


\begin{figure}[t]
    \centering
{\includegraphics[width=0.41\textwidth]{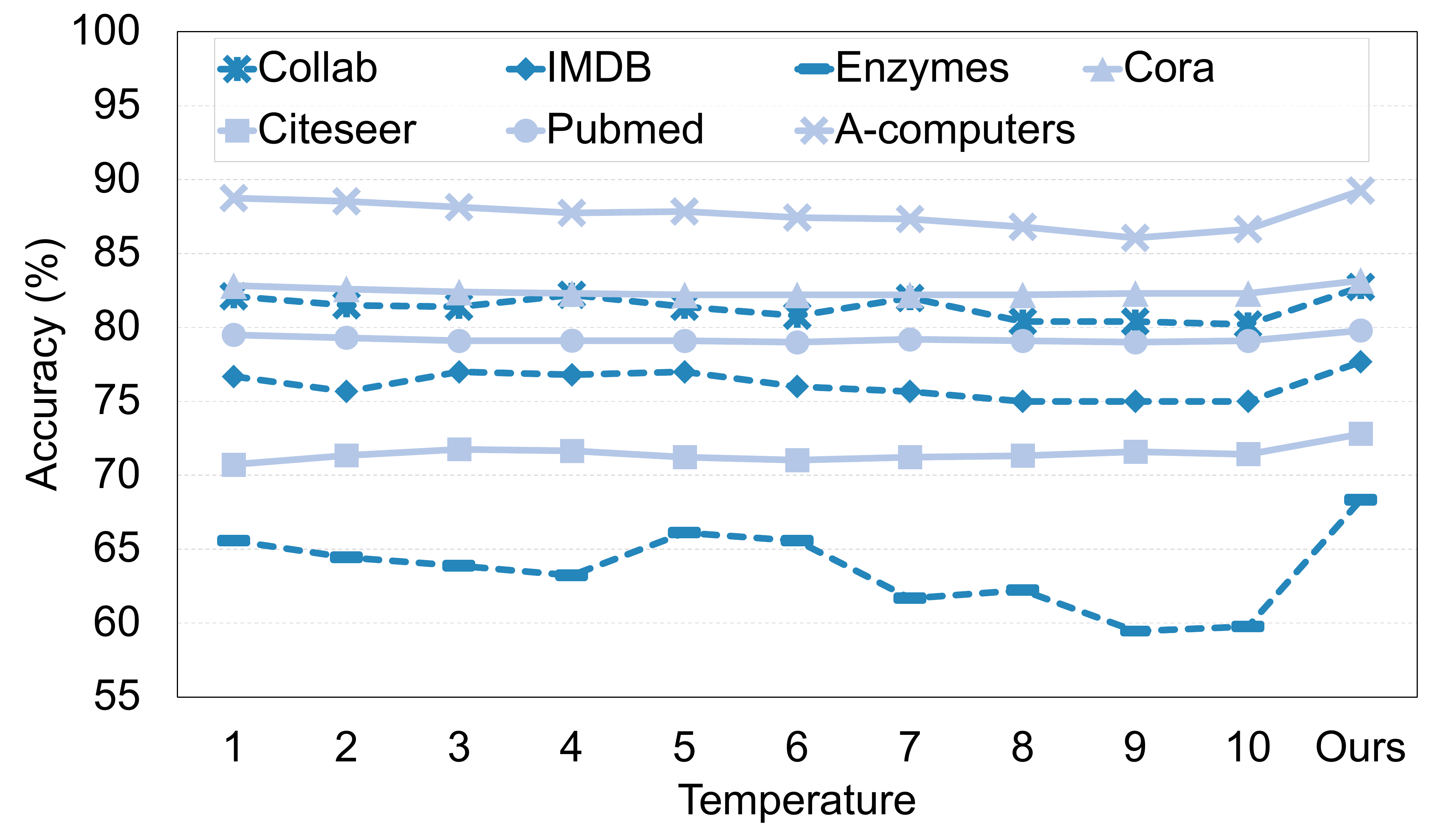}\label{fig:tempe}}
    \caption{Results of BGNN with adaptive temperature and fixed temperatures from 1 to 10 (without weight boosting).}
    \label{fig:ablation_temp}
\end{figure}

\begin{figure}[t]
    \centering
    {\includegraphics[ width=0.41\textwidth]{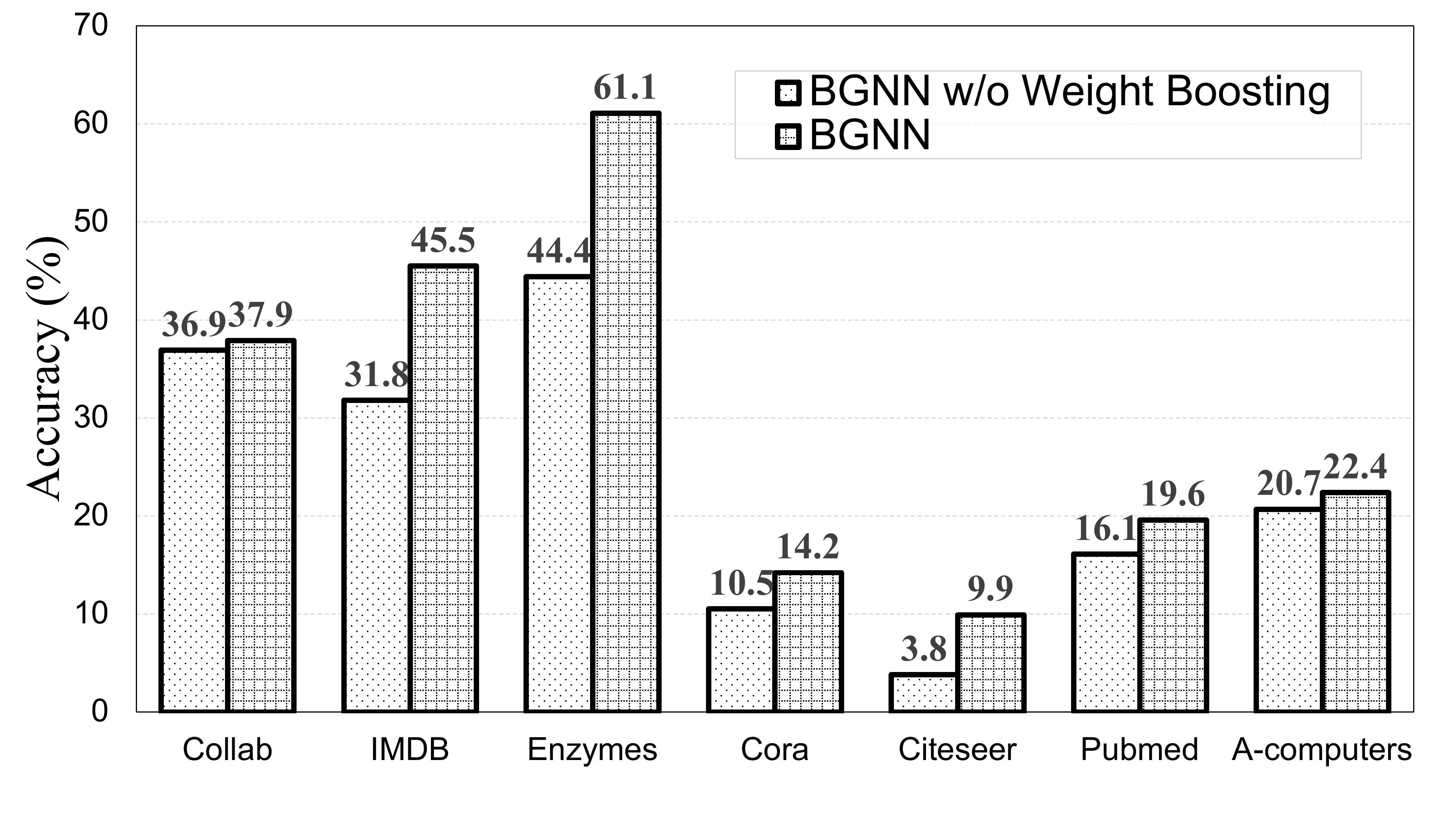}\label{fig:wrong}}
    \caption{\textit{Right:} Results of BGNN without weight boosting and BGNN on the mis-classifed nodes by the teacher GNN.}
    \label{fig:ablation_boosting}
\end{figure}


\subsection{How Does the Selection of Teacher GNNs Influence the Student GNNs?} 
We study the performance of the student GNNs when they are taught by different teachers. The results are shown in Figure~\ref{fig:teach_student}. In 18 out of the 21 settings, the student learning from different architectures outperforms the respective ones learning from the same teacher GNNs. One exception is that, on Enzymes, the student GCN learning from GCN is more powerful than the GCNs learning from other GNNs. But for all the other datasets, we still find that GCN learns more from other GNNs. If we only consider the node classification task (the four datasets on the right part of each histogram), it becomes more obvious that a student may learn more from different architectures. In 11 out of the 12 settings, the students learning from the same architecture have the worst performance. However, for the multi-teacher setting, we do not find a clear clue of how to decide the order of teachers. As shown in Table~\ref{tab:baseline_multi}, BGNN(m) usually outperforms BGNN(s), but it seems arbitrary when a specific order will be better. 

\subsection{Ablation Study}
\label{sec:ablation}

\noindent\textbf{Does the adaptive temperature matter for knowledge distillation?} 
In Table~\ref{tab:ablation}, we can see that the original BGNN performs better than its variation without the adaptive temperature (replacing the adaptive temperature with a fixed temperature). It means the adaptive temperature indeed helps BGNN learn more knowledge from the teacher GNN than the fixed temperature. Then, we further compare the adaptive temperature with the fixed temperatures from 1 to 10. Note that we do not train the models with the weight boosting to exclude its impact. In the left of Figure~\ref{fig:ablation_temp}, we find that the adaptive temperature achieves the highest accuracy for all the seven datasets. This means that the adaptive temperature can make the best trade-off between the dark knowledge and true label information to maximize the transferred knowledge.

\noindent\textbf{Does the weight boosting help?} In Table~\ref{tab:ablation}, we observe that BGNN performs better than BGNN without the weight boosting, which confirms its effectiveness of weight boosting. We argue that the power originates from the improvement on the nodes mis-classified by the teacher GNN. This is verified by the right figure of Figure~\ref{fig:ablation_boosting}, which compares the performance with and without weight boosting on the mis-classified nodes. We see that BGNN obtains higher accuracy with weight boosting, because the weight boosting forces the student to pay more attention on these mis-classified nodes.



\section{Conclusion}
In this paper, we propose a novel KD framework to boost a single GNN by combining various knowledge from different GNNs. We develop a sequential KD training strategy to merge all knowledge. To transfer more useful knowledge from the teacher model to the student model, we propose an adaptive temperature for each node based on the teacher's confidence. Additionally, our boosting weight module helps the student to pick up the knowledge missed by the teacher GNN. The effectiveness of our methods has been proved on both node classification and graph classification tasks. 

\section*{Acknowledgements}
This work was supported by National Science Foundation under the NSF Center for Computer Assisted Synthesis (C-CAS), grant number CHE-2202693. We thank all anonymous reviewers for their valuable comments and suggestions. 


\bibliography{ref}

\newpage
\clearpage
\appendix
\section{Derivation of Equation (2) in the Paper}
In this section, we elaborate on the detailed derivation of Equation~\eqref{equation:anal} in the paper. As the softened logit vector produced by the teacher $\hat{\pmb{y}}_v^\mathrm{T}$ is independent to $z_{v,c}$:
\begin{equation}
    -\frac{\partial}{\partial z_{v,c}} \sum_{j=1}^{C} \hat{y}_{v,j}^\mathrm{T}\mathrm{log}(\hat{y}_{v,j}^\mathrm{S}) = - \sum_{j=1}^{C}\hat{y}_{v,j}^\mathrm{T} \cdot\frac{\partial}{\partial z_{v,c}} \mathrm{log}(\hat{y}_{v,j}^\mathrm{S}).
    \label{equation:a}
\end{equation}
Then we first expand the term $\mathrm{log}(\hat{y}_{v,j}^\mathrm{S})$ into:
\begin{align}
    \mathrm{log}(\hat{y}_{v,j}^\mathrm{S}) & = \mathrm{log}(\frac{e^{z_{v,j}/\tau_v}}{\sum_{k=1}^{C}e^{z_{v,k}/\tau_v}}) \nonumber \\ & = z_{v,j}/\tau_v - \mathrm{log}(\sum_{k=1}^{C}e^{z_{v,k}/\tau_v}).
\end{align}
Based on the above result, the partial derivative of Equation~\eqref{equation:a} becomes:
\begin{equation}
    \frac{\partial}{\partial z_{v,c}} \mathrm{log}(\hat{y}_{v,j}^\mathrm{S}) = \frac{\partial z_{v,j}/\tau_v}{\partial z_{v,c}} - \frac{\partial}{\partial z_{v,c}}\mathrm{log}(\sum_{k=1}^{C}e^{z_{v,k}/\tau_v}),
    \label{equation:b}
\end{equation}
where the first term on the right hand side is:
\begin{equation}
    \frac{\partial z_{v,j}/\tau_v}{\partial z_{v,c}}=\left\{
    \begin{array}{rcl}
        \frac{1}{\tau_v} &, & \text{if}\ j = c  \\
        0 & , & \text{otherwise}
    \end{array}
    \right. .
    \label{equation:c}
\end{equation}
We can concisely rewrite Equation~\eqref{equation:c} using the indicator function: $\frac{1}{\tau_v}\cdot 1\{\cdot\}$: if the argument is true,  the indicator function returns $1$ and the result of Equation~\eqref{equation:c} becomes $\frac{1}{\tau_v}$; otherwise, the result is $0$ in both cases. Applying this indicator function and the chain rule, we can rewrite Equation~\eqref{equation:b} as:
\begin{align}
     \frac{\partial}{\partial z_{v,c}} \mathrm{log}(\hat{y}_{v,j}^\mathrm{S}) \nonumber \\  
     = \frac{1}{\tau_v} \cdot 1 \{j=c\}
     & - \frac{1}{\sum_{k=1}^{C}e^{z_{v,k}/\tau_v}} \cdot \frac{\partial}{\partial z_{v,c}} (\sum_{k=1}^{C}e^{z_{v,k}/\tau_v}).
     \label{equation:d}
\end{align}
Next, we can derive the partial derivative of the summation term in Equation~\eqref{equation:d} as follows:
\begin{align}
    & \frac{\partial}{\partial z_{v,c}} (\sum_{k=1}^{C}e^{z_{v,k}/\tau_v}) \nonumber \\ 
    & = \frac{\partial}{\partial z_{v,c}}(e^{z_{v,1}/\tau_v} + e^{z_{v,2}/\tau_v} + \cdots +  e^{z_{v,C}/\tau_v})\nonumber\\ &= \frac{\partial }{\partial z_{v,c}}(e^{z_{v,c}/\tau_v}) = \frac{1}{\tau_v}(e^{z_{v,c}/\tau_v}).
\end{align}
Plugging this result into Equation~\eqref{equation:d}, we have:
\begin{align}
     \frac{\partial}{\partial z_{v,c}} \mathrm{log}(\hat{y}_{v,j}^\mathrm{S}) & = \frac{1}{\tau_v} \cdot 1 \{j=c\} - \frac{1}{\tau_v}\frac{e^{z_{v,c}/\tau_v}}{\sum_{k=1}^{C}e^{z_{v,k}/\tau_v}} \nonumber \\  & = \frac{1}{\tau_v} \cdot 1 \{j=c\} - \frac{1}{\tau_v} \hat{y}_{v,c}^\mathrm{S}.
\end{align}
Therefore, the derivative of the KD loss, shown in Equation~\eqref{equation:a}, can be written as:
\begin{align}
     & -\frac{\partial}{\partial z_{v,c}}  \sum_{j=1}^{C} \hat{y}_{v,j}^\mathrm{T}\mathrm{log}(\hat{y}_{v,j}^\mathrm{S}) \nonumber \\
     & = -\sum_{j=1}^{C}\hat{y}_{v,j}^\mathrm{T} \cdot (\frac{1}{\tau_v} \cdot 1 \{j=c\} - \frac{1}{\tau_v} \hat{y}_{v,c}^\mathrm{S})  \\ 
     &= \frac{1}{\tau_v}\sum_{j=1}^{C}\hat{y}_{v,j}^\mathrm{T}\hat{y}_{v,c}^\mathrm{S} -\sum_{j=1}^{C}\hat{y}_{v,j}^\mathrm{T} \cdot \frac{1}{\tau_v} \cdot 1\{j=c\},
     \label{equation:e}
\end{align}
where $\sum_{j=1}^{C}\hat{y}_{v,j}^\mathrm{T} = 1$. Since the value of the indicator function is $1$ only when $j=c$, the second term of the right hand side in Equation~\eqref{equation:e} becomes $\frac{1}{\tau_v} \hat{y}_{v,c}^\mathrm{T}$. Thus, Equation~\eqref{equation:e} can be further simplified as:
\begin{align}
     -\frac{\partial}{\partial z_{v,c}} \sum_{j=1}^{C} \hat{y}_{v,j}^\mathrm{T}\mathrm{log}(\hat{y}_{v,j}^\mathrm{S}) & = \frac{1}{\tau_v}\sum_{j=1}^{C}\hat{y}_{v,j}^\mathrm{T}\hat{y}_{v,c}^\mathrm{S} - \frac{1}{\tau_v}\hat{y}_{v,c}^\mathrm{T} \nonumber\\ & = \frac{1}{\tau_v} \hat{y}_{v,c}^\mathrm{S} - \frac{1}{\tau_v} \hat{y}_{v,c}^\mathrm{T},
     \label{equation:f}
\end{align}
which is exactly the derivation of Equation~\eqref{equation:anal} in the paper: $\frac{\partial{\mathcal{L}_{KD}}}{\partial{z_{v,c}}} = \frac{1}{\tau_v} (\hat{y}_{v,c}^\mathrm{S} - \hat{y}_{v,c}^\mathrm{T})$.

Note that the above derivation is used for the cross entropy between the prediction of teacher $\hat{y}_{v,c}^\mathrm{T}$ and that of the student $\hat{y}_{v,c}^\mathrm{S}$. When the supervised loss between the ground-truth $y_v$ and the prediction of student $\hat{y}_{v,c}^\mathrm{S}$ is considered, it is sufficient to simply substitute $\hat{y}_{v,c}^\mathrm{T}$ in Equation~\eqref{equation:f} with $y_v$. This result has been used directly in the Pre-analysis paragraphs of Section~\ref{sec: temp} in the paper.

\section{Dataset Details}
The dataset statistics for graph classification and node classification can be found in Table~\ref{tab:dataset}. In the following, we introduce the detailed description of the datasets. 

\begin{table}
    \centering
        \setlength{\tabcolsep}{1mm}{
        \scalebox{0.8}{
        \begin{tabular}{>{\small}c|>{\small}c|>{\small}c|>{\small}c|>{\small}c|>{\small}c|>{\small}c}
        \toprule
        Task & Dataset & \# Graphs & \# Nodes & \# Edges & \# Features & \# Classes \\
        \midrule
        \multirow{3}{*}{Graph} & \textsc{Enzymes} & 600 & $\sim$32.6 & $\sim$124.3 & 3 & 6 \\
        & \textsc{IMDB-Binary} & 1,000 & $\sim$19.8 & $\sim$193.1 & - & 2\\
        & \textsc{Collab} & 5,000 & $\sim$74.5 & $\sim$4,914.4 & - & 3\\
        \midrule
        \multirow{4}{*}{Node}& \textsc{Cora} & - & 2,485 & 5,069 & 1,433 & 7\\
        & \textsc{Citeseer} & - & 2,110 & 3,668 & 3,703 & 6\\
        & \textsc{Pubmed} & - & 19,717 & 44,324 & 500 & 3\\
        & \textsc{A-computers} & - & 13,381 & 245,778 & 767 & 10\\
        \bottomrule
    \end{tabular}
    }
    }
    \caption{Dataset statistics.}
    \label{tab:dataset}
\end{table}
\subsection{Graph Classification Datasets}
\textbf{Collab} is a scientific collaboration dataset, incorporating three public collaboration datasets~\cite{leskovec2005graphs}, namely High Energy Physics, Condensed Matter Physics, and Astro Physics. Each graph represents the ego-network of scholars in a research field. The graphs are classified into three classes based on the fields. We randomly split this dataset in our experiments. Since there is no node feature for this dataset, we provide synthetic node features using the one-hot degree transforms\footnote{https://pytorch-geometric.readthedocs.io/en/latest/\\modules/datasets.html\#torch\_geometric.datasets.TUDataset}. 

\noindent\textbf{IMDB~\cite{yanardag2015deep}} is a movie collaboration dataset. Each graph represents the ego-network of an actor/actress, where each node represents one actor/actress and each edge indicates the co-appearance of two actors/actresses in a movie. The graphs are classified into two classes based on the genres of the movies: Action and Romance. We randomly split the dataset in our experiments. Same as the Collab dataset, we provide synthetic node features using the one-hot degree transforms.

\noindent\textbf{Enzymes~\cite{borgwardt2005protein}} is a benchmark graph dataset. Each graph represents a protein tertiary structure. The node features are categorical labels. The graphs are classified based on six EC top-level classes. We randomly split the dataset in our experiments.

\subsection{Node Classification Datasets}
\textbf{Cora~\cite{sen2008collective}} is a benchmark citation dataset. Each node represents a machine-learning paper and each edge represents the citation relationship between two papers. The nodes are associated with sparse bag-of-words feature vectors. Papers are classified into seven classes based on the research fields. We use the standard fixed splits for this dataset.

\noindent\textbf{Citeseer~\cite{sen2008collective}} is another benchmark citation dataset, where each node represents a computer science paper. It has a similar configuration to Cora, but it has six classes and larger features for each node. We also use the standard fixed splits for this dataset.

\noindent\textbf{Pubmed~\cite{namata2012query}} is a citation dataset as well, where the papers are related to diabetes from the PubMed dataset. The node features are TF/IDF-weighted word frequencies. Nodes are classified into three classes based on the types of diabetes addressed in the paper. This dataset provides standard fixed splits, which we use in our experiments.

\noindent\textbf{A-computers~\cite{shchur2018pitfalls}} is extracted from Amazon co-purchase graph~\cite{mcauley2015image}, where each node represents a product and each edge indicates that the two products are frequently bought together. The node features are bag-of-words encoded product reviews. Products are classified into ten classes based on the product category. We randomly split this dataset in our experiments, as this dataset does not come with a standard fixed split.


\section{Additional Experimental Results}
\begin{figure}
    \centering
    \vspace{-0.15in}
    \includegraphics[width=0.4\textwidth]{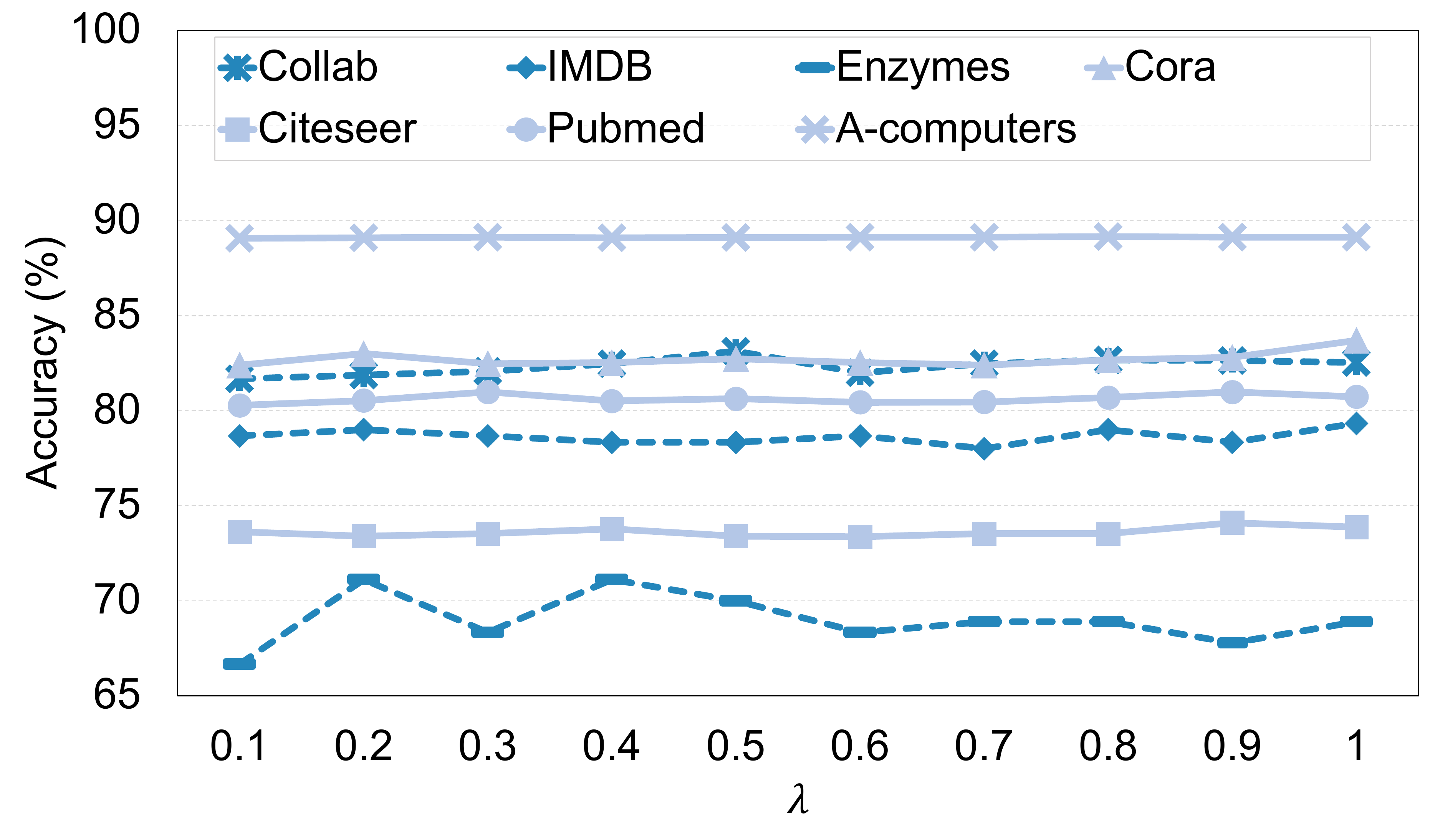}
    \vspace{-0.1in}
    \caption{Results of BGNN with different $\lambda$ on all the seven datasets. $\lambda$ varies from 0.1 to 1.}
    \label{fig:lambda}
\end{figure}

\noindent\textbf{What's the influence of $\lambda$? } We conduct experiments with GAT as the teacher and GCN as the student. Figure~\ref{fig:lambda} shows that the accuracy of BGNN is stable for most datasets except Enzymes, when the trade-off factor $\lambda$ varies from 0.1 to 1. It indicates that our method is insensitive to $\lambda$ in general.

\section{Implementation Details}
Both our method BGNN and other baselines are implemented using PyTorch. Specifically, we use the PyG~\cite{Fey/Lenssen/2019} library for GNN algorithms, and Adam~\cite{kingma2014adam} for optimization with weight decay set to $5 \times 10^{-4}$. Other training settings vary across datasets, which are described as follows.

\noindent\textbf{BGNN Training Details.} We use 2-layer GCN~\cite{kipf2017semi}, GraphSage~\cite{hamilton2017inductive} and GAT~\cite{velickovic2018graph} with activation and dropout operation in intermediate layers as backbone GNNs. We use ReLU as the activation function for GCN and GraphSage, and use ELU activation~\cite{clevert2015fast} for GAT. The hidden dimensions of different GNNs are selected per dataset to match the performance reported in the original publication under supervised settings. For GCN and GraphSage, we use the same configuration of hidden dimensions: 16 on Cora and Pubmed; 32 on IMDB, Enzymes, and Citeseer; 64 on Collab; and 128 on A-computers. For GAT, we build the first layer with 8 heads of 8-dimension on most datasets, except 4 heads of 8-dimension on collab and 8 heads of 16-dimension on Enzymes and A-computers.

We take mini-batch to train BGNN on graph classification datasets for memory efficiency. The batch size is 32 on all three datasets. We further add a batch normalization operation in the intermediate layers of GNNs for better prediction performance. We use full-batch on node classification datasets. What's more, to obtain relatively stable results on graph classification tasks, we take the top 5 results over 10 rounds as the results for the corresponding tasks, which is applied on both BGNN and baselines. For the teacher and student GNNs, we set the number of layers and the hidden dimension of each layer to the same numbers as the corresponding supervised settings. The supervised settings are described in the previous paragraph. For BGNN, we conduct the hyperparameter search of the weights for $\lambda$ from [0.1, 0.5, 1, 5, 10], and the learning rate from [0.005, 0.01, 0.05].


\noindent\textbf{Baseline Details.} For BAN~\cite{furlanello2018born}, LMTN~\cite{you2017learning} and KD~\cite{hinton2015distilling}, we select the supervisedly trained GNNs as teachers in the same way as for BGNN. For Ensemble, we take average logits of three supervisedly trained GNNs for both node classification task and graph classification task. GNNs used in the baselines share the same architectures as those in BGNN.

\noindent\textbf{Hardware Details.} We run all experiments on a single NVIDIA P100 GPU with 16GB RAM, except the experiments using GAT~\cite{velickovic2018graph} on Collab are run on a single NVIDIA V100 GPU with 32GB RAM.


\end{document}